\documentclass[10pt,twocolumn,letterpaper]{article}

\usepackage{iccv}
\usepackage{times}
\usepackage{epsfig}
\usepackage{graphicx}
\usepackage{amsmath}
\usepackage{amssymb}
\usepackage{booktabs}
\usepackage{relsize}
\usepackage{tabularx}
\usepackage{multirow}
\usepackage{multicol}
\usepackage{pifont}
\usepackage{comment}
\newcommand{\cmark}{\ding{51}}%
\newcommand{\xmark}{\ding{55}}%
\usepackage[dvipsnames]{xcolor}

\usepackage[breaklinks=true,bookmarks=false]{hyperref}
\usepackage[capitalize]{cleveref}

\iccvfinalcopy 

\def\iccvPaperID{2876} 

\ificcvfinal\fi

\newcommand{\onoff}[1]{}
\definecolor{brandoncolor}{RGB}{0, 0, 0}
\definecolor{rohitcolor}{RGB}{0, 0, 0}
\definecolor{sidcolor}{RGB}{0, 0, 0}
\definecolor{ganeshcolor}{RGB}{0, 0, 0}

\newcommand{\brandonedit}[1]{\textcolor{brandoncolor}{#1}}

\newcommand{\sidedit}[1]{\textcolor{sidcolor}{#1}}
\newcommand{\ganeshedit}[1]{\textcolor{ganeshcolor}{#1}}

\newcommand{\bD}{\mathbf{D}}
\newcommand{\bbeta}{\boldsymbol{\beta}}
\newcommand{\btheta}{\boldsymbol{\theta}}
\newcommand{\bR}{\boldsymbol{R}}
\newcommand{\bt}{\boldsymbol{t}}
\newcommand{\bM}{\boldsymbol{M}}
\newcommand{\bm}{\boldsymbol{m}}
\newcommand{\bI}{\boldsymbol{I}}
\newcommand{\bS}{\boldsymbol{S}}
\newcommand{\bk}{\boldsymbol{k}}

\newcolumntype{Y}{>{\centering\arraybackslash}X}

\begin{document}

\title{Mesh Strikes Back: Fast and Efficient Human Reconstruction from RGB videos}

\author{
Rohit Jena\textsuperscript{1}\thanks{Work done during an internship at Amazon} 
\and Pratik Chaudhari\textsuperscript{1}
\and James C. Gee\textsuperscript{1} 
\and Ganesh Iyer\textsuperscript{2}
\and Siddharth Choudhary\textsuperscript{2}
\and Brandon M. Smith\textsuperscript{2} \\
{\centering 
\textsuperscript{1}University of Pennsylvania \hspace{1cm} \textsuperscript{2}Amazon.com, Inc} 
}

\maketitle
\ificcvfinal\thispagestyle{empty}\fi

\begin{abstract}
Human 
reconstruction and synthesis
from monocular RGB videos is a challenging problem due to clothing, occlusion, texture discontinuities and sharpness, and frame-specific pose changes.
Many methods employ deferred rendering, NeRFs and implicit methods to represent clothed humans, on the premise that mesh-based representations cannot capture complex clothing and textures from RGB, silhouettes, and keypoints alone.
We provide a counter viewpoint to this fundamental premise by optimizing a SMPL+D mesh and an \brandonedit{efficient, multi-resolution} texture representation 
\brandonedit{using only} RGB images, binary silhouettes and sparse 2D keypoints.
\brandonedit{Experimental results demonstrate that our approach is more capable of capturing geometric details compared to visual hull, mesh-based methods.} 
\brandonedit{We show competitive novel view synthesis and improvements in novel pose synthesis compared to NeRF-based methods, which introduce noticeable, unwanted artifacts.}
By restricting the solution space to the SMPL+D model combined with differentiable rendering, we obtain dramatic speedups in compute, training \brandonedit{times (up to 24x)} and inference \brandonedit{times (up to 192x)}.
Our method therefore can be used as is or as a fast initialization to NeRF-based methods.

\end{abstract}
\section{Introduction}
\label{sec:intro}
Our goal is to generate detailed, personalized, and animatable 3D human models. 
This has many downstream applications, including teleconferencing, entertainment, surveillance, and realistic synthetic data generation. 
3D body scanners are the gold standard when it comes to 3D reconstructions. 
Results are accurate and realistic, but scanners tend to be expensive and ungainly. 
Moreover, they require additional postprocessing, \brandonedit{such as registration and rigging,} before they can be used.
More recently, CV-based systems have demonstrated recovering realistic 3D human geometry and appearance from monocular images or videos.
In the case of monocular \textit{images}, a predictor can be learned from a dataset of real or synthetic humans \cite{Sengupta20bmvc,Xiu22icon}.
However, this is ill-conditioned because a single view is insufficient to estimate the entire 3D human geometry or appearance completely or accurately.
Extending it to multi-view or 360$^\circ$ video input would require 
running inference 
for all frames
and fusing per-frame texture and mesh information.
%
3D geometry and texture recovery is therefore formulated as an optimization problem.
This is an alternative to expensive 3D scanning and motion capture pipelines.

\begin{figure}[t!]
    \centering
    \includegraphics[width=0.9\linewidth,height=0.8\linewidth]{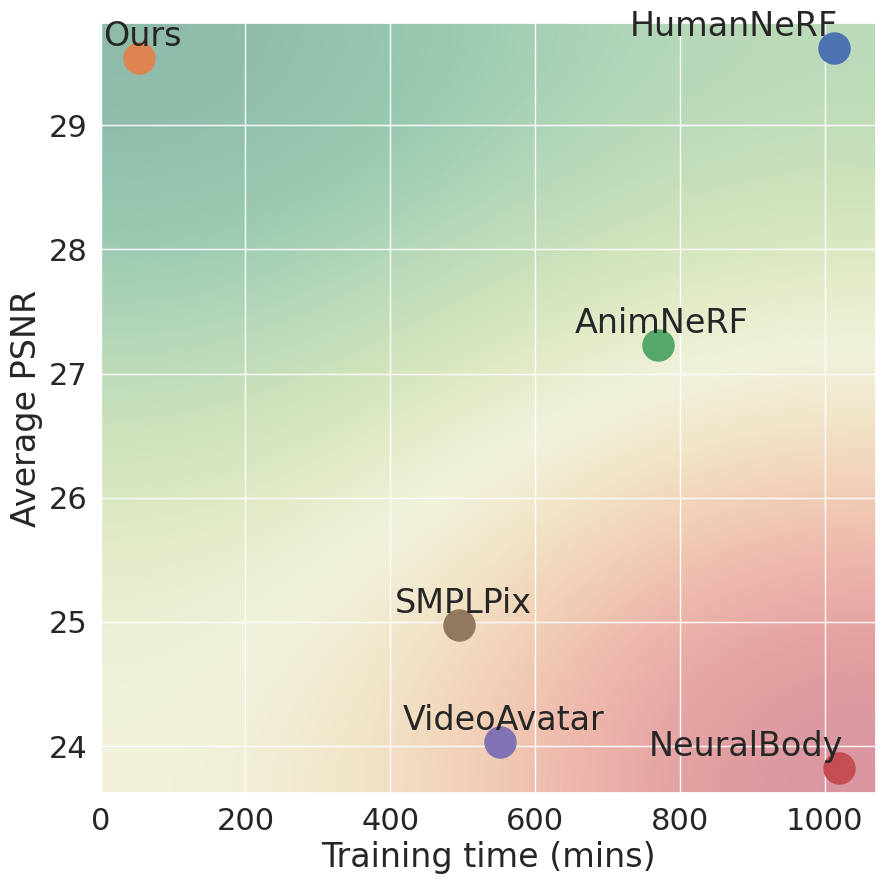}
    \caption{\textbf{Performance vs training time tradeoff}: Our method maximizes the performance to training time tradeoff compared to other methods employing meshes, NeRFs and deferred rendering.}
    \label{fig:perf}
\end{figure}

Human-specific NeRFs have recently become popular under this latter, video-based formulation.
There are two downsides to {typical} NeRFs: (1) the highly unconstrained solution space of NeRFs and the use of volume rendering leads to long training and rendering times \eg, {up to a few days for training~\cite{chen:arxiv2021:animner,weng:cvpr2022:humannerf,Liu21neuralactor} and up to minutes for rendering 
images~\cite{hedman:iccv2021:bakingnerf}}, and (2) dense viewpoints are required.
On the contrary, mesh-based reconstruction methods can be faster and less compute-intensive because they do not optimize over a volume. 
The solution space of the mesh is highly constrained due to flexible yet accurate parameterized human body priors like SMPL~\cite{SMPL:2015}. 
Moreover, mesh-based methods may perform better than NeRFs under sparser viewpoints without expensive pretraining or strong regularization that misses geometric details (~\cite{Niemeyer_2022_CVPR}), due to the human shape prior. 

\brandonedit{However, mesh-based methods that rely on a visual hull or silhouette-based approach (\eg, \cite{alldieck:cvpr2018:peoplesnap}) suffer from ambiguous shape recovery, \ie, any concave surface is not captured by the visual hull} and thus cannot be recovered.
We show that silhouette-based optimization is highly ill-posed (Sec.~\ref{sec:toyprob}). 
To overcome this ambiguity, recent methods augment silhouettes with depths or normals~\cite{Xiu22icon, dong:cvpr2022:pina}.
However, obtaining ground truth depths or normals 
is expensive, and prediction can be error-prone.
RGB images provides additional information \brandonedit{that} can be used across multiple frames to better recover 3D information. 
However, directly employing an RGB loss is non-trivial because of \brandonedit{differentiability challenges} in rasterization.
NeRFs use the inherent `softness' of the occupancy field (in the volume integral) to reason about self-occluded portions of the body, allowing pose-refinement~\cite{chen:arxiv2021:animner}.
We use a soft differentiable rendering pipeline~\cite{liu2019softras} to emulate this behavior.
This allows us to use `analysis-by-synthesis` as part of our optimization problem.
We adopt an approach that uses texture information as part of the optimization similar to NeRFs, but we parameterize the body similar to mesh-based reconstruction methods.
Meshes are simpler and more efficient, which affords significant speedup (up to 24x in training time and 192x in inference time) compared to NeRFs.

However, a naive mesh-based optimization to simultaneously minimize RGB and silhouette losses does not work because of \textit{moving targets}.
The problem of \textit{moving targets} occur when the RGB losses between the image and a partially learnt texture representation hinder the mesh deformation and vice versa.
We propose a method to reduce the ill-posedness and mitigate the problem of moving targets in optimization using a two-stage optimization. 
We demonstrate that our 3D reconstruction results are similar in quality and accuracy to NeRFs, and significantly better than existing mesh-based reconstruction methods.
\sidedit{In addition, we show competitive results in novel view and novel pose synthesis} 
\brandonedit{compared to NeRF-based methods.}

In summary, this paper makes three main contributions:
(1) To our knowledge, we present the first method to incorporate photogrammetric losses in the context of generating human avatars from monocular videos using a mesh representation (Sec.~\ref{sec:method}). This allows us to optimize texture and geometry using \textit{analysis-by-synthesis} in our optimization without additional auxiliary inputs.
(2) An efficient, multi-resolution texture representation using hash encoding capable of capturing fine details is proposed.
Unlike texel-based representations, capacity is not wasted on uniform-texture regions.
(3) To mitigate the moving targets problem, where partially learnt texture and geometry hinder each other's loss functions, a novel two-stage optimization is proposed. This ensures stability and optimal convergence.

\section{Related Work}
\label{sec:related}


\subsection{Human reconstruction via prediction}
Recovering 3D human shape and pose estimation using a parametric 3D body model 
\brandonedit{such as} SCAPE~\cite{Pishchulin17scape}, SMPL~\cite{Loper15tog}, SMPL-X~\cite{Pavlakos19smplx}, STAR~\cite{Osman20STAR} or GHUM~\cite{Xu21ghum} is an active area of research in the 
\brandonedit{CV community}.
Most of these approaches directly predict model parameters using a learned model~\cite{Kanazawa18cvpr,Kocabas20cvpr,Sengupta21iccv,Sengupta20bmvc,Sengupta21bmvc,Sengupta21cvpr,Smith193dv,Varol18bodynet,Yu21iccv,Yu22cvpr}.
Recent approaches (\eg,~\cite{Kolotouros19iccv,Smith193dv,Sengupta20bmvc,Sengupta21cvpr}) 
have improved on prior methods by directly regressing body shape. 
However, these approaches can only represent the shape and pose of a minimally clothed body and fail to model complex topology due to clothing, hair, \brandonedit{\etc.} 
BodyNet~\cite{Varol18bodynet} and DeepHuman~\cite{Zheng19DeepHuman} attempt to predict volumetric representations of the human model from a single image.
Implicit representations are an interesting alternative for representing high-fidelity 3D geometry without requiring the entire output volume be kept in memory.
In contrast to explicit representations, implicit representations define a surface as a level set of a function.
Recent methods, such as PiFu, PiFuHD and PHORHUM~\cite{Saito19pifu,Saito20pifuhd,Alldieck22phorhum}, learn an implicit surface representation estimated based on pixel aligned features and the depth of 3D locations.
Some recent methods combine the benefits of both explicit and implicit representations to represent clothed people~\cite{Corona22lvd,bhatnagar:eccv2020:ipnet,Zheng20pamir,Cao22jiff,He21ARCHAC,Xiu22icon}.
These methods require 3D ground-truth supervision, limiting their \brandonedit{applicability} to a few datasets and their \brandonedit{ability to generalize beyond in-distribution poses.} 
Recently, implicit representations have been used to learn a generative model of 3D people in clothing \cite{Chen22gdna,Alldieck20imghum,Deng19NASA,Saito21SCANAnimate,Chen21SNARF}.
However, these approaches require ground truth posed 3D meshes or RGB-D video sequence to learn a model
\cite{Tao21thuman,Zheng19DeepHuman,Patel21agora,twindom,renderpeople,axyz}. 
All of the methods share a common limitation, \ie, predicting the human shape and texture from a single image is ill-posed, and incorrectly regressed predictions are not iteratively refined using 
auxiliary signals.

\subsection{Human reconstruction via optimization}
\textbf{Recovering pose}: Some of the earliest approaches fit model parameters via optimization at test time~\cite{alldieck:cvpr2018:peoplesnap,Bogo16eccv,Kolotouros21iccv}.
Bogo \etal~\cite{Bogo16eccv} optimize SMPL parameters by minimizing the joint reprojection loss and prior terms, whereas~\cite{Kolotouros21iccv,Kolotouros19iccv,Zhang21iccv} 
employ a likelihood term over the pose via a learned network, and perform iterative regression to minimize reprojection or multiview losses.
Pavlakos \etal~\cite{pavlakos:2022:recon3d} reconstruct SMPL parameters of multiple humans from videos of TV shows.
However, these methods only recover a pose with generic shapes and do not capture subject-specific deformations and textures.

\textbf{Recovering shape and texture}: Alldieck \etal~\cite{alldieck:cvpr2018:peoplesnap} 
extended this approach to monocular video by fusing the unposed silhouettes from all frames to generate a consensus mesh. Each human silhouette, extracted from a video frame, defines a constraint on the human shape, which can be used to estimate deviations in shape.
The main problem with this method is that shape is ambiguous, \ie, concave surfaces are not captured. 
One would need to use auxiliary inputs like depths or normals to disambiguate the problem \cite{Xiu22icon,alldieck:3dv2018:humanmono}, but depth prediction systems can introduce their own errors. 

Mildenhall~\etal~\cite{Mildenhall20nerf,Xie22neuralfields} pioneered NeRFs for representing static scenes with a color and density field without requiring any 3D ground truth supervision. 
Recently this approach has been extended to reconstruct clothed humans as well \cite{peng:cvpr2021:neuralbod,Su21anerf,chen:arxiv2021:animner,weng:cvpr2022:humannerf,jiang:eccv2022:neuman,Weng20vid2actor,Liu21neuralactor,wang:eccv2022:arah,Jiang22selfrecon,yao:arxiv2022:monoclothedbody,Te22neuralcapture}. 
These approaches use SMPL as a prior to unpose the human body across multiple frames by transforming the rays from observation space to canonical space which is then rendered using a NeRF.
PINA~\cite{dong:cvpr2022:pina} learns a SDF and a learned deformation field to create an animatable avatar from an RGB-D image sequence.
Chen~\etal~\cite{Chen22mobilenerf} used a polygon rasterization pipeline to speed up NeRF rendering as a post-process; 
however, their approach does not reduce NeRF training times.
\sidedit{Some recent methods have improved the efficiency of scene-agnostic NeRFs (\eg~\cite{yu2021plenoctrees,yu_and_fridovichkeil22plenoxels,mueller2022instantngp,Garbin21FastNeRF,Reiser21kilonerf})}.
These methods demonstrate fast training and rendering times, but adapting them to include an animatable human shape prior requires expensive nearest neighbor operations that erode efficiency gains. 
However, most of these approaches are computationally expensive.
In addition, the resulting representation may have poor generalization to OOD poses.

\begin{table*}[t!]
    \centering
    \small
    \begin{tabularx}{0.97\linewidth}{Xccccccc} \toprule
    \textbf{Method} & \textbf{RGB loss} & \textbf{Mask loss} & \textbf{KPS loss} & \textbf{Representation} & \textbf{Novel pose} & \textbf{Training time} & \textbf{GPU} \\
    \hline
     VideoAvatar~\cite{alldieck:cvpr2018:peoplesnap} & \xmark & \cmark & \cmark & SMPL+D & \cmark  & 16 hours & NA \\
    AnimNeRF~\cite{chen:arxiv2021:animner} & \cmark & \cmark & \xmark & NeRF & \cmark & 26 hours & 48GB \\
    Neuralbody~\cite{peng:cvpr2021:neuralbod} & \cmark & \cmark & \xmark & NeRF & \xmark & 14 hours & 48GB \\
    HumanNeRF~\cite{weng:cvpr2022:humannerf} & \cmark & \cmark & \xmark & NeRF & \cmark & 72 hours & 48GB \\
    NeuralActor~\cite{Liu21neuralactor} & \cmark & \cmark & \xmark & NeRF & \cmark & 48 hours & 256GB \\
    SCARF~\cite{yao:arxiv2022:monoclothedbody} & \cmark & \cmark & \xmark & NeRF+SMPLX-D & \cmark & 40 hours & 32GB \\
    \hline
        Ours & \cmark & \cmark & \cmark & SMPL+D & \cmark & \textbf{$<$1hour} & \textbf{5GB} \\
    \bottomrule
    \end{tabularx}
    \caption{Comparison of different methods for human reconstruction. Our simple yet clever use of NeRF-like losses with an SMPL+D representation bridges the gap between mesh-based and NeRF-based optimization with dramatic speedups and compute savings.}
    \label{tab:compare}
\end{table*}

Our method falls in the category of shape and texture optimization, being most similar to~\cite{alldieck:cvpr2018:peoplesnap,chen:arxiv2021:animner,Jiang22selfrecon,weng:cvpr2022:humannerf}, where we aim to contrast NeRFs with its mesh-based equivalent formulation.
In contrast to NeRFs, an explicit mesh representation combined with a carefully chosen optimization scheme enables our method to recover accurate geometry, while ensuring photo-realistic rendering, at substantially lower computational and time costs.
We show that, contrary to conventional wisdom, our method, when employed with the correct optimization objective, can recover complex geometry (like loose clothing, skirts, long hair, hoodies, \etc).
Because we constrain the solution space using a mesh, our method is computationally inexpensive, and can be run on a single consumer-grade GPU in about an hour.
Our representation allows us to use differentiable rasterization pipelines, drastically reducing its training and inference time.
Note that our goal is not to replace NeRF methods, which have asymptotically better performance due to more
representational capacity.
\brandonedit{Rather, \emph{we provide a computationally inexpensive alternative that can be used for real-time rendering and/or on-device applications, or to bootstrap NeRF optimization.}}
Unlike~\cite{peng:cvpr2021:neuralbod,Liu21neuralactor}, which have reported failure cases for \brandonedit{out-of-distribution} poses, our method's accuracy is only limited by the artefacts of the skinning function.
A comparison with relevant methods is provided in Fig.~\ref{tab:compare}.
\begin{figure}
    \centering
    \includegraphics[width=\linewidth]{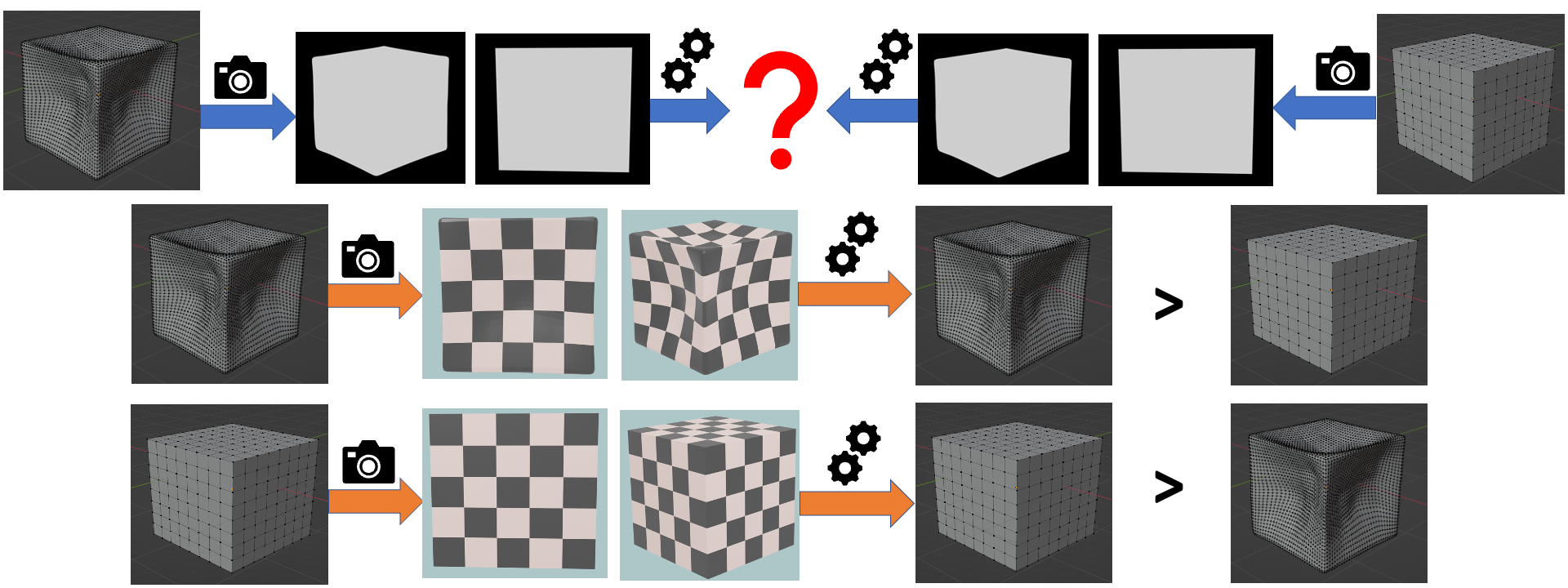}
    \caption{
    A toy example hightlighting that methods based on visual hulls alone cannot recover concavities in the underlying object.
    Optimization from visual hull is ill-conditioned, which is disambiguated by multi-view RGB consistency.
    This idea is used in NeRFs, and in this work we show that it is possible 
    to use RGB and visual hull to optimize a mesh representation.
    }
    \label{fig:cube}
\end{figure}

\section{Method}
\label{sec:method}
This paper focuses on jointly recovering accurate geometry and realistic textures from a monocular RGB video of a person using a mesh representation.
First, we illustrate the ambiguity in mesh reconstruction from visual hull only, and how multi-view RGB consistency can help disambiguate the problem (Section~\ref{sec:toyprob}).
This allows us to avoid using auxiliary inputs, such as depths, landmarks, and normals, which are either expensive to obtain or error-prone if predicted.
Next, we discuss the shape and texture representations (Sec~\ref{sec:shape_model},~\ref{sec:texnet}) used in our optimization.
To enable photogrammetric losses to backpropagate to a parameterized mesh representation, we describe the differentiable rendering pipeline (Section~\ref{sec:diffrender}).
Finally, we describe the forward model (Section~\ref{sec:forward_model}), loss functions (Section~\ref{sec:loss_functions}) and training pipeline (Section~\ref{sec:two_stage_training}) to jointly recover the clothed human geometry and texture of the subject.
\subsection{A motivating toy problem}
\label{sec:toyprob}
We use the following toy problem 
to illustrate the ambiguity of using visual hulls for mesh reconstruction. 
Consider a cube and the same cube but with all its faces dented inwards, shown in Fig.~\ref{fig:cube} (top).
Rendering the visual hull of both objects gives us the same set of binary silhouettes for all camera angles, making it ambiguous for any optimization scheme to recover a unique mesh from the set of silhouettes.
This ambiguity necessitates the use of auxiliary inputs, \eg, depth or normals \cite{dong:cvpr2022:pina,Xiu22icon,embodiedhands}. 
Now consider the same scenario, but with the same overlaid texture on both objects.
In this case, the two objects have different renders in Fig.~\ref{fig:cube} (middle \& bottom) 
from the same viewpoints,
disambiguating the shape of the underlying object when optimized with a multi-view RGB consistency framework.
This is the idea used in NeRFs \cite{Mildenhall20nerf} to recover the occupancy and radiance volume from images alone. 
Therefore, one can use multi-view RGB consistency as a surrogate to depth, normals, \etc.
The key to using RGB images for mesh optimization is to assign a unique RGB value to each point on the mesh, 
such that it can guide the mesh vertices to produce consistent renderings in all views. 
However, doing so introduces a `moving target' problem (partially optimized mesh and RGB hinder each other's learning) which is non-trivial to optimize.
Empirically, we find that carefully formulating the optimization  problem (Sec~\ref{sec:two_stage_training}) allows us to capture complex geometry, \brandonedit{including} hoodies, loose shirts, pants, skirts, and \brandonedit{voluminous} hair, better than prior work that uses visual hulls alone for optimization.  


\subsection{Geometry model}
\label{sec:shape_model}
Optimizing a high-fidelity textured avatar from monocular or multi-camera RGB video requires us to learn geometry and corresponding texture on the learned geometry.
Methods using neural rendering learn a canonical or conditional volume from scratch without using any structural priors \cite{Mildenhall20nerf}.
Instead of learning a volume from scratch, we use the parametric SMPL human model ~\cite{Loper15tog}.
We learn per-frame pose and camera parameters $\{(\btheta_i, \bR_i, \bt_i)\}_{i \in \{1..n\}}$ and a common shape parameter $\bbeta$.
Since the shape parameter is a low-dimensional embedding capturing human shape, we also learn a per-vertex offset matrix $\bD \in \mathbb{R}^{V\times3}$ where $V$ is the number of vertices in the mesh.
Unlike ~\cite{alldieck:3dv2018:humanmono} we show that the base SMPL+D model is enough to capture most geometric details (Fig.~\ref{fig:geom} and in Appendix) and we do not need to use a subdivided SMPL model.


\subsection{Texture representation}
\label{sec:texnet}

\begin{figure}[ht!]
    \centering
    \includegraphics[width=0.8\linewidth]{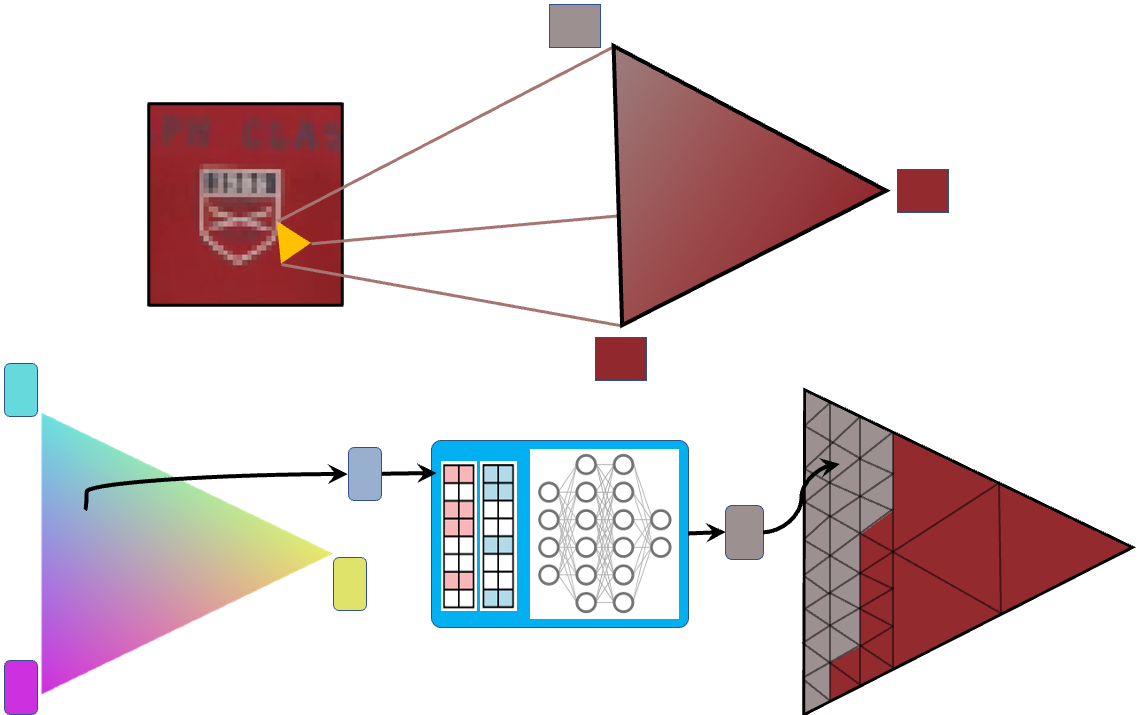}
    \caption{(\textbf{Top}) Traditional texturing methods use fixed per-vertex UV coordinates and linearly interpolate the color from the UV coordinates over the face leading to blurry texture. 
    (\textbf{Bottom}) We use learnt per-vertex 3D texture coordinates (in cyan,magenta,yellow) and linearly interpolate the texture coordinates over the face.
    The interpolated texture coordinate is input into the multi-res hash encoding \cite{mueller2022instantngp} which allows us to represent sharp textures within the face (illustrated by how the network treats the input space across multiple resolutions).
    }
    \label{fig:texnet-illus}
\end{figure}

Mesh-based representations generally use predefined UV coordinates for each vertex \cite{alldieck:cvpr2018:peoplesnap,alldieck:3dv2018:humanmono}, and the RGB value at any point on the face is determined by evaluating the UV coordinate of the point, and using interpolation from the RGB values in the UV map. 
This is converted into a texel-map ~\cite{kato2018renderer,liu2019softras}.
This representation, although simple, is not adaptive to the 
\brandonedit{complexity} of texture on the mesh.
Meshes may have regions of low or high texture (\eg, \brandonedit{solid-color} clothing \brandonedit{versus} faces, hair, \brandonedit{or patterned} clothing) which may require a more flexible texture representation.
Bilinear interpolation also contributes to blurry textures
, making it unsuitable for \brandonedit{high-frequency or discontinuous} 
textures.
Moreover, deferred rendering ~\cite{smplpix,thies2019deferred} may overfit to the UV distribution of the training frames and may not generalize to UV distributions of OOD poses (Fig.~\ref{fig:novelpose}).

To alleviate these problems, we take inspiration from \brandonedit{M\"uller \etal~\cite{mueller2022instantngp} who proposed} a multi-resolution hash encoding of the input to capture high-frequency details.
The main idea of this work is to use implicit hash collisions to average the gradients at a particular hash lookup entry, and therefore weigh the representation appropriately.
We use the same idea to learn a high frequency texture representation.
Human avatars have 
\brandonedit{varying levels of texture complexity} across their surface---regions such as faces, logos on clothing, \etc have high levels of detail, \brandonedit{whereas} 
regions \brandonedit{such as} 
skin and \brandonedit{solid-color} clothing have low levels of detail, and \brandonedit{require little} 
representation capacity to learn.
The hash-encoding representation would learn this implicitly by finding the best embedding which leads to the 
\ganeshedit{least} RGB reconstruction loss, effectively `splitting' its representation capacity between high and low-frequency details.
%
For a given frame $i$ using shape $\bbeta$, body pose $\btheta_i$ and deformation $\bD$ we produce a mesh $\mathcal{\bM}_i$. 
This mesh is then projected to the image plane using camera extrinsics $(\bR_i, \bt_i)$ and a fixed intrinsics matrix using the projective transformation to mesh $\mathcal{\bm}_i$.
We perform rasterization on the projected mesh to output an image containing face indices and barycentric coordinates of the face index for each pixel.
Let face $f_j$ be rendered at pixel $p$ with barycentric coordinates $\{ w_{j1}, w_{j2}, w_{j3} \}$ such that $w_{jk} \ge 0 $
and $\sum_{k=1}^3 w_{jk} = 1$.
If the texture coordinates of $f_j$ are given by $\{ t_{j1}, t_{j2}, t_{j3}\}$, then the rendered point on the face has the texture coordinate $t = \sum_{k=1}^3 w_{jk} t_{jk}$.
%
%
The texture coordinate is a low-dimensional input that is passed into the hash encoder and MLP to output the RGB color at pixel $p$.
Conventionally, hardcoded 2D texture coordinates are used for vertices of each face, which are interpolated and used to lookup \brandonedit{values in} 
a UV map (typically an image).
Due to UV unwrapping of a closed mesh, some mesh vertices have multiple texture coordinates.
In contrast, we use a learned 3D texture coordinate for each vertex.
This serves two purposes: (1) it sidesteps the need for UV unwrapping by moving the textures coordinates into 3D space, and (2) learnable coordinates allow us to expand or shrink the 3D coordinates of each vertex, which in turn accommodate different levels of detail for each face (See Appendix). 



\subsection{Differentiable rendering}
\label{sec:diffrender}
A \brandonedit{major} 
advantage of using NeRF-like methods is that the volume rendering process is fully differentiable.
Moreover, the gradients with respect to the integral (to compute the color) takes occlusions into account.
This allows NeRFs to learn a robust occupancy volume and RGB colors to minimize rendering errors.
In contrast, rasterization is an inherently non-differentiable operation with respect to the vertices of the mesh ~\cite{liu2019softras}.
\brandonedit{Many solutions have been proposed} 
to approximate gradients of the vertices from gradients in rendered images
\brandonedit{(\eg, \cite{kato2018renderer, liu2019softras, loper:eccv2014:opendr}).}
We use SoftRas ~\cite{liu2019softras} due to 
its ability to flow gradients to the occluded and far-range vertices, allowing us to perform pose refinement via analysis-by-synthesis.
Complex pose changes such as bringing an occluded limb into view, rotating joints, etc. can now be performed since SoftRas allows us to pass gradients into the occluded parts of the render. 
Empirically, we observe that SoftRas is helpful in updating body pose when a \brandonedit{small amount} of joint rotation is required. 
To learn the texture, we need the exact forward rasterization to map texture coordinates to RGB values. 
In SoftRas, we can set the softening parameters $\gamma = \sigma = 0$ and use the RGB loss to guide the texture learning.
However, this leads to numerical instability and rendering artefacts in SoftRas.
To mitigate this issue, we use NMR ~\cite{kato2018renderer} to propagate texture gradients.


\subsection{Forward model}
\label{sec:forward_model}
In this section, we describe the forward model for a given frame $i$. 
The SMPL+D and camera parameters $(\bbeta, \btheta_i, \bD, \bR_i, \bt_i)$ are used to generate the mesh $\bM_i$ which is projected into the image plane as $\bm_i$.
The projected mesh and texture network parameters are passed into NMR and SoftRas to give us
an opaque and translucent RGB image respectively. 
For texture parameters $\phi$, we have:
\begin{equation}
    \left[\hat{\bI}_{i,\text{NMR}}; \hat{\bI}_{i,\text{SR}}\right] = \text{NMR}(\bm_i; \phi), \text{SoftRas}(\bm_i; \text{sg}(\phi),\sigma,\gamma)
\end{equation}
where $\sigma,\gamma$ are the face blur and depth scale parameters of SoftRas, and $\text{sg}$ is the stop-grad operator.

\subsection{Loss functions}
\label{sec:loss_functions}
\brandonedit{In this section, we describe the losses used to generate our results.}
Let the masked ground-truth image for frame $i$ be $\bI_i$ and ground-truth binary foreground be $\bS_i$.
\vspace*{-5pt}
\paragraph{Image losses.}
The RGB losses are given by:
\begin{equation}
    \mathcal{L}_{i,\text{RGB}} = \| \bI_i - \hat{\bI}_{i,\text{NMR}} \|_1 + \| \bI_i - \hat{\bI}_{i,\text{SR}} \|_1 
\end{equation}
We also obtain a binary silhouette using NMR and projected mesh $\bm_i$, which we denote as $\hat{\bS}_i$.
The silhouette loss is defined as \brandonedit{an IOU loss}:
\begin{equation}
    \mathcal{L}_{i, \text{Sil}} = 1 - \frac{\sum_p \hat{\bS}_i(p)\cdot \bS_i(p)}{\sum_p (\hat{\bS}_i(p) + \bS_i(p) - \hat{\bS}_i(p)\cdot \bS_i(p))}.
\end{equation}

\paragraph{Keypoint loss.}
\brandonedit{We add a keypoint loss to help guide the pose of the SMPL model. 
For simplicity,} we use only keypoints corresponding to limbs (elbows, wrists, knees, ankles) and nose.
From the mesh $\bM_i$, we project keypoints $\hat{\bk}_i$ and \brandonedit{encourage their proximity to target 2D keypoints $\bk_i$ via the loss:}
\begin{equation}
    \mathcal{L}_{i,\text{kps}} = \| \hat{\bk}_i - \bk_i \|_2^2.
\end{equation}
\brandonedit{We run HRNet~\cite{ke:cvpr2019:hrnet} on each frame to produce $\bk_i$ for people-snapshot and use the given keypoints for ZJU Mocap.}
\begin{table*}[ht!]
    \centering
    \small
    \setlength{\tabcolsep}{5pt}
    \begin{tabularx}{\linewidth}{Xccccc|ccccc|ccccc}
    \toprule 
    \multicolumn{16}{c}{\textbf{Novel view (people-snapshot)}} \\ \hline
    \multirow{2}{*}{\textbf{Subject ID}} & \multicolumn{5}{c|}{\textbf{PSNR}$\uparrow$} & \multicolumn{5}{c|}{\textbf{SSIM}(x100) $\uparrow$} & \multicolumn{5}{c}{\textbf{LPIPS}(x100) $\downarrow$} \\
    & \textbf{VA} & \textbf{SP} & \textbf{NB} & \textbf{AN} & \textbf{Ours} & \textbf{VA} & \textbf{SP} & \textbf{NB} & \textbf{AN} & \textbf{Ours} & \textbf{VA} & \textbf{SP} & \textbf{NB} & \textbf{AN} & \textbf{Ours}  \\ \hline
    \textbf{m3c} & 22.91 & 22.94 & 23.98 & \textbf{29.43} & \underline{29.40} & 93.16 & 92.56 & 96.12 & \textbf{97.11} & \underline{96.24} & 4.87 & 6.89 & 7.24 & \textbf{1.85} & \underline{2.65} \\
    \textbf{m4c} & 22.63 & 21.43 & 22.84 & \textbf{27.50} & \underline{26.31} & 93.22 & 92.66 & \underline{94.81} & \textbf{95.87} & 94.27 & 6.00 & 8.04 & 10.93 & \textbf{3.77} & \underline{5.30} \\
    \textbf{f3c} & 22.10 & 21.80 & \underline{23.19} & 22.96 & \textbf{27.25} & 94.35 & 93.95 & \underline{95.83} & 94.56 & \textbf{96.21} & 5.43 & 5.61 & 10.22 & \underline{4.61} & \textbf{3.47} \\
    \textbf{f4c} & 23.49 & 22.64 & 22.18 & \underline{29.03} & \textbf{29.61} & 93.99 & 93.27 & 95.63 & \textbf{96.88} & \underline{96.61} & 4.12 & 5.92 & 8.52 & \textbf{2.10} & \underline{2.42} \\
    \hline
    \multicolumn{16}{c}{\textbf{Novel view (ZJU Mocap)}} \\ \hline
    & \textbf{VA} & \textbf{SP} & \textbf{NB} & \textbf{HN} & \textbf{Ours} & \textbf{VA} & \textbf{SP} & \textbf{NB} & \textbf{HN} & \textbf{Ours} & \textbf{VA} & \textbf{SP} & \textbf{NB} & \textbf{HN} & \textbf{Ours}  \\ \hline
    \textbf{C377}  & 24.48 & 27.28 & 24.81 & \textbf{30.86} & \underline{30.58} & 93.12 & 94.64 & \underline{97.17} & \textbf{97.45} & 96.51 & 9.01 & 5.82 & 5.71 & \textbf{2.58} & \underline{4.48} \\
    \textbf{C386}  & 27.67 & 29.22 & 25.08 & \textbf{33.36} & \underline{33.28} & 93.72 & 95.80 & \underline{97.27} & \textbf{97.29} & 96.22 & 7.98 & 10.65 & 4.86 & \textbf{3.39} & \underline{4.18} \\
    \textbf{C387}  & 23.30 & 24.27 & 23.60 & \textbf{28.58} & \underline{28.07} & 92.58 & 95.08 & \underline{96.08} & \textbf{96.10} & 94.73 & 9.18 & 12.08 & 6.88 & \textbf{3.96} & \underline{6.40} \\
    \textbf{C392}  & 25.70 & 28.66 & 24.35 & \textbf{31.42} & \underline{31.35} & 92.89 & 95.64 & \textbf{96.86} & \underline{96.83} & 96.05 & 9.52 & 7.54 & 6.37 & \textbf{3.76} & \underline{5.37} \\
    \textbf{C393}  & 23.45 & 24.83 & 24.17 & \textbf{28.89} & \underline{28.35} & 92.30 & 93.60 & \textbf{96.35} & \underline{95.83} & 93.88 & 10.99 & 12.77 & 6.66 & \textbf{4.22} & \underline{6.43} \\
    \textbf{C394}  & 24.46 & 27.34 & 23.97 & \underline{30.73} & \textbf{31.21} & 91.67 & 95.58 & \textbf{96.43} & \underline{96.16} & 95.57 & 11.28 & 9.07 & 6.71 & \textbf{3.75} & \underline{4.91} \\ \hline
    \multicolumn{16}{c}{\textbf{Novel pose (ZJU Mocap)}} \\ \hline
    \textbf{C377}  & 24.36 & 27.00 & 23.84 & \textbf{30.50} & \underline{30.48} & 93.25 & 96.51 & \underline{96.78} & \textbf{97.41} & 96.54 & 8.29 & 5.81 & 5.59 & \textbf{2.69} & \underline{4.27} \\
    \textbf{C386}  & 28.34 & 30.38 & 23.26 & \underline{33.55} & \textbf{34.03} & 93.84 & 96.60 & \underline{96.46} & \textbf{97.20} & 96.16 & 7.43 & 9.79 & 5.50 & \textbf{3.41} & \underline{4.00} \\
    \textbf{C387}  & 23.02 & 23.80 & 23.15 & \textbf{29.02} & \underline{28.43} & 92.83 & 95.38 & \underline{95.58} & \textbf{96.44} & 95.23 & 8.46 & 11.47 & 6.77 & \textbf{3.25} & \underline{5.71} \\
    \textbf{C392}  & 25.83 & 29.12 & 22.46 & \underline{31.43} & \textbf{32.22} & 92.98 & \underline{96.44} & 95.97 & \textbf{96.89} & 96.37 & 9.40 & 7.26 & 7.03 & \textbf{3.70} & \underline{5.01} \\
    \textbf{C393}  & 23.50 & 24.79 & 22.41 & \textbf{29.32} & \underline{28.62} & 92.49 & 95.07 & \underline{95.45} & \textbf{96.09} & 94.07 & 10.58 & 12.65 & 7.13 & \textbf{3.86} & \underline{6.47} \\
    \textbf{C394}  & 24.33 & 26.99 & 22.19 & \underline{30.20} & \textbf{30.36} & 91.72 & 95.64 & \underline{95.43} & \textbf{95.95} & 95.10 & 11.09 & 8.44 & 7.29 & \textbf{4.07} & \underline{5.25} \\
    \bottomrule
    \end{tabularx}
    \caption{Results on novel view and pose synthesis on people-snapshot and ZJU datasets. Best results are in \textbf{bold} and 2nd best is \underline{underlined}.}
    \label{tab:novelview}
\end{table*}

\paragraph{Mesh regularization losses.}
We add some regularization to the mesh representation.
\brandonedit{This is especially important for the free-form per-vertex offsets $\bD$.}
Unlike previous works (\eg, \cite{alldieck:cvpr2018:peoplesnap}), we observe that imposing a low-deformation loss reduces performance (Sec~\ref{sec:geomrecon}). 
This is because loose clothing need not necessarily correspond to a low deformation from the underlying SMPL model.
We only encourage normal consistency of adjacent faces in the mesh.
Let $\bM_D$ be the mesh generated from the SMPL parameters $(\bbeta, \boldsymbol{0}, \bD, \mathbb{I}, \boldsymbol{0})$
\brandonedit{and} let $\bM_0$ be generated from the SMPL parameters $(\bbeta, \boldsymbol{0}, \boldsymbol{0}, \mathbb{I}, \boldsymbol{0})$.
Let $f_j$ be the $j^{\text{th}}$ face of $\bM_D$ and $f'_j$ be the $j^{\text{th}}$ face of $\bM_0$.
With some abuse of notation, for two faces $f_j$ and $f_k$, we denote $|f_j \cap f_k|$ as the number of vertices that are shared between both faces.
The normal consistency loss is then given as:
\begin{equation}
    \mathcal{L}_{NC} = \mathlarger{\sum}_{|f_j \cap f_k| = 2} (1 - \hat{n}_{f_i}\cdot \hat{n}_{f_j}),
\end{equation}
where $\hat{n}_{f}$ is the outward normal of face $f$.
We also encourage each face in the mesh to have the same area with and without the deformation.
This respects the relative sizes of faces corresponding to different regions in the mesh.
If $A_{f}$ represents the unsigned area of a face $f$, the face area loss is given as:
\begin{equation}
    \small
    \mathcal{L}_{FA} = \mathlarger{\sum}_j \Bigg( \frac{A_{f_j}}{A_{f'_j}} + \frac{A_{f'_j}}{A_{f_j}}\Bigg).
\end{equation}
We prefer this loss instead of the L2 loss $\| A_{f_j} - A_{f'_j} \|_2^2$ because the gradients of L2 loss
are small when the area $A_{f_j}$ approaches 0.
On the other hand, we want to penalize shrinkage or expansion equally. The loss we propose is of the form $x + \frac{1}{x}$ which achieves its minima at $x = 1$ for positive $x$. 
The total loss is: 
\begin{multline}
    \mathcal{L}_f = \sum_i \left( \lambda_{\text{RGB}}\mathcal{L}_{i,\text{RGB}} + \lambda_{\text{Sil}}\mathcal{L}_{i,\text{Sil}} +  \lambda_{\text{kps}}\mathcal{L}_{i,\text{kps}} \right) \\
    + \lambda_{NC}\mathcal{L}_{NC} + \lambda_{FA}\mathcal{L}_{FA}
\end{multline}

\brandonedit{Unlike \cite{chen:arxiv2021:animner},} 
we do not add any other regularization terms on $\bbeta$ or temporal pose consistency or deviation terms. 

\subsection{Two-stage training}
\label{sec:two_stage_training}
Given a forward model to render RGB and silhouettes from mesh parameters $\bbeta, \btheta_i, \bD, \bR_i, \bt_i$ and texture parameters $\phi$, an intuitive way to learn all the parameters is to jointly optimize them.
Let $\Theta = \{\phi^*, \bbeta^*, \bD^*, \{\btheta_i^*, \bR_i^*, \bt_i^*\}_{i \in \{1\ldots n\}} \}$ be the set of all optimizable parameters. The optimization is of the form:
$
    \Theta^* = \arg\min_\Theta \mathcal{L}_f.
$
This optimization is still highly underconstrained. 
Meshes obtained using this procedure are jagged \brandonedit{with blurry textures} because 
texture and deformation parameters locally optimize their own parameters (examples in Appendix) leading to the \textit{moving target} problem.
To alleviate this problem, we propose a two-stage training procedure.
In the first stage, we use a \textit{per-face} RGB value as a `base' texture color instead of the full texture network.
This allows \brandonedit{for} 
a coarse alignment of the mesh vertices for each frame.
Constraining the color of each face to just one optimizes the mesh parameters to place it in the best possible location and scale to minimize RGB and silhouette losses, thus ensuring 
photogrammetric
consistency in the optimization.
In the second stage, the deformations, shape, and per-face RGB values are fixed, and the texture network is trained with per-frame pose \brandonedit{refinement.}
This \brandonedit{allows for fine-grained alignment of the poses for each frame.} 
Implementation details are \brandonedit{provided} 
in the Appendix.

\begin{figure}
    \centering
    \includegraphics[width=0.95\linewidth]{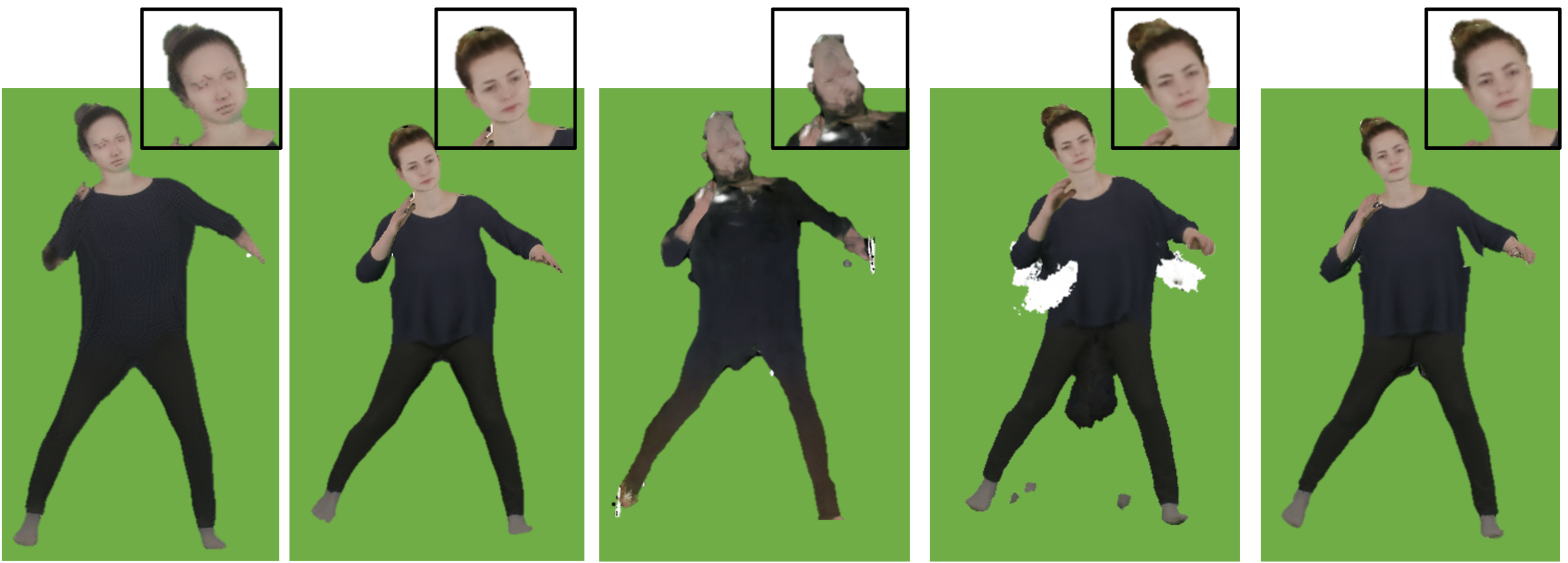}
    \includegraphics[width=0.95\linewidth]{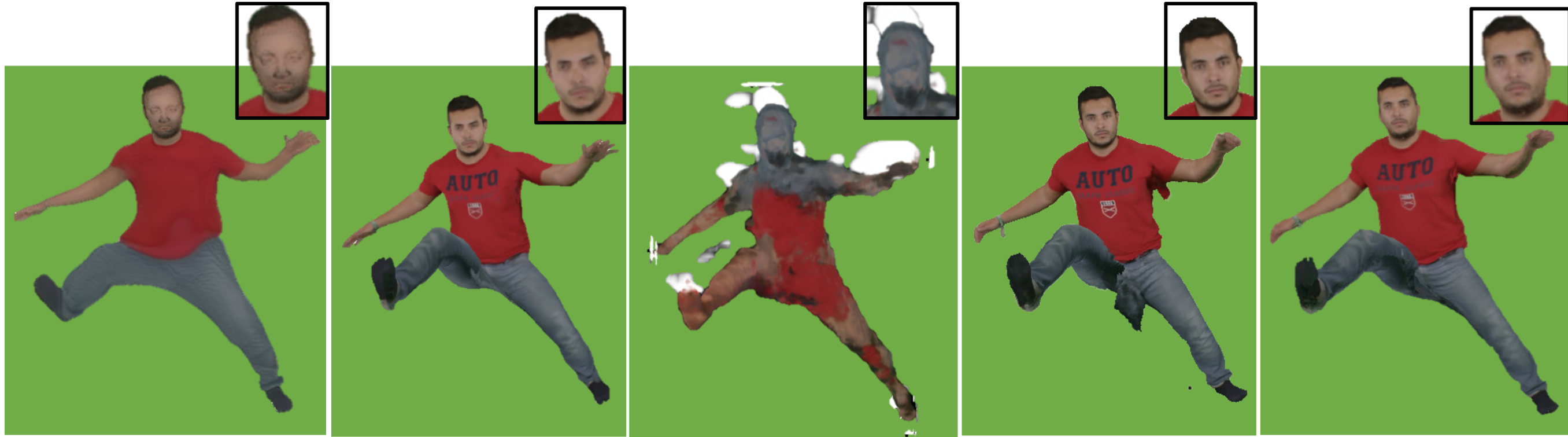}
    \caption{Novel pose synthesis (on \textbf{m3c} and \textbf{f3c}) from left to right: SMPLpix, VideoAvatar, NeuralBody, AnimNeRF, Ours. NeuralBody does not generalize to novel poses, AnimNeRF introduces cloud and tearing artifacts. Our method preserves detailed texture and doesn't produce artifacts. More results in Appendix.}
    \label{fig:novelpose}
\end{figure}

\section{Results}
\label{sec:results}
To evaluate our method, we look at the following aspects of geometry and texture recovery: (1) novel-view and pose synthesis, (2) training/inference time and compute requirements, (3) geometry reconstruction.
For experiments (1-2), we use People Snapshot ~\cite{alldieck:cvpr2018:peoplesnap} and ZJU Mocap ~\cite{peng:cvpr2021:neuralbod} datasets.
For (3), we use the Self-Recon synthetic dataset ~\cite{Jiang22selfrecon}.
For people-snapshot, we follow the same subjects and experiment setup as ~\cite{chen:arxiv2021:animner}, and for ZJU Mocap we use the same set of subjects as ~\cite{weng:cvpr2022:humannerf}.
For ZJU Mocap, we use frames 0-450 in cameras 1,7,13,19 for training, and the rest of the frames for novel pose reconstruction.
We use frames 0-450 from cameras 5,10,15,20 for novel view reconstruction.
We choose baselines across a spectrum of representation choices: {S}MPL{P}ix ~\cite{smplpix} (\textbf{SP}) which uses deferred rendering, {V}ideo{A}vatar ~\cite{alldieck:cvpr2018:peoplesnap} (\textbf{VA}) which performs SMPL+D optimization, {N}eural{B}ody~\cite{peng:cvpr2021:neuralbod} (\textbf{NB}), {H}uman{N}eRF~\cite{weng:cvpr2022:humannerf} (\textbf{HN}) and {A}nim{N}eRF~\cite{chen:arxiv2021:animner} (\textbf{AN}) which are SOTA NeRF methods.


\begin{figure}
    \centering
    \includegraphics[width=0.9\linewidth]{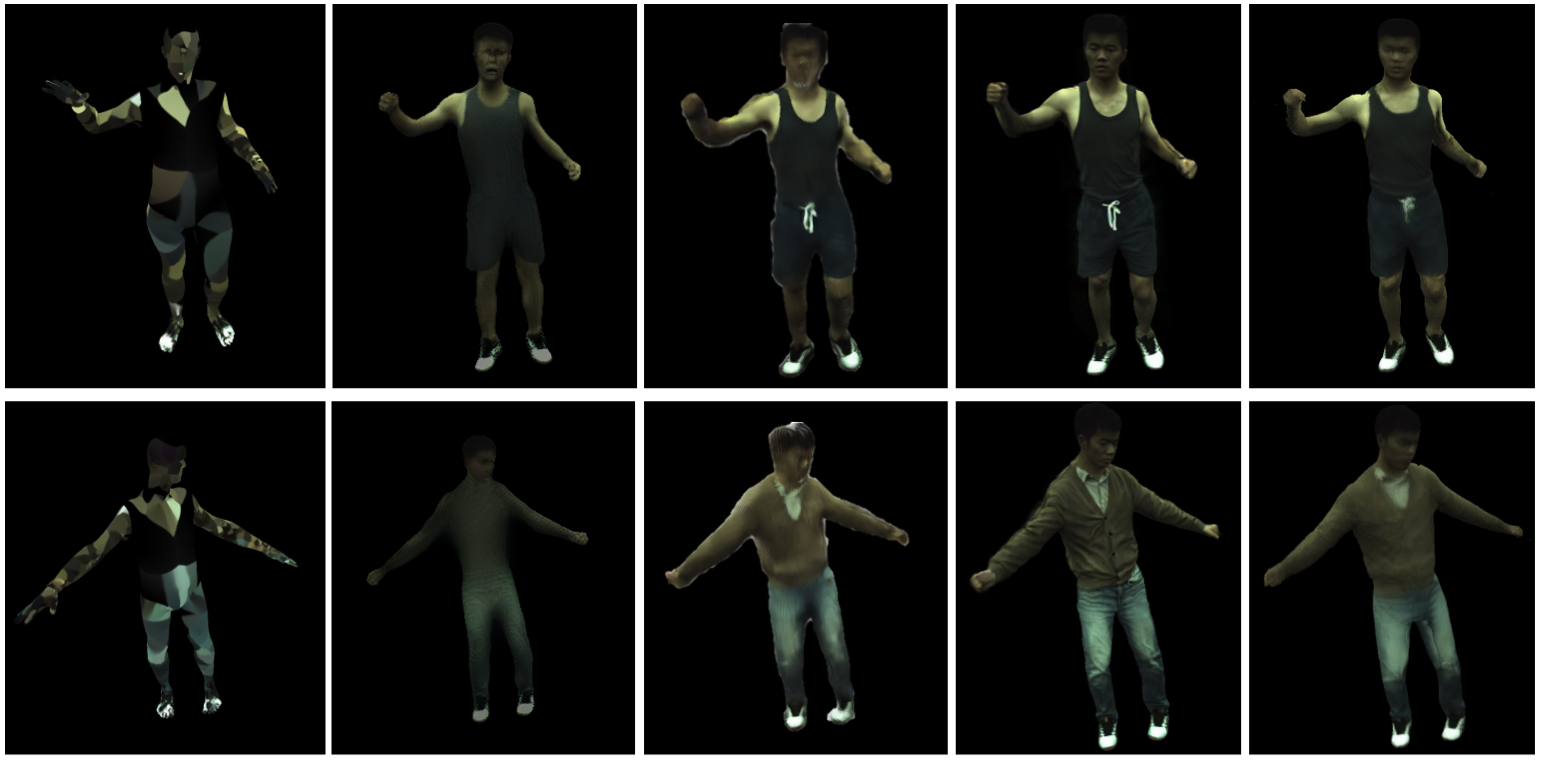}
    \caption{\textbf{Novel view on ZJU dataset, Left to right}: VideoAvatar, SMPLPix, NeuralBody, HumanNeRF, Ours. Our method performs dramatically better than VA/SMPLPix and is comparable to NeRF methods at significant training and inference speedups.}
    \label{fig:zju}
\end{figure}

\subsection{Novel view synthesis}
We evaluate novel view synthesis
by holding back a certain set of test frames of the subject to check the quality of reconstruction for those frames (similar to ~\cite{chen:arxiv2021:animner}).
For ZJU Mocap, we evaluate frames 0-450 from cameras 5,10,15,20.
Results are shown in Table \ref{tab:novelview}.
Our method performs very competitively with AnimNeRF and HumanNeRF in terms all three metrics (PSNR, SSIM, LPIPS), and outperforms all other baselines by a substantial margin.
On ZJU Mocap, our method competes with HumanNeRF consistently on the PSNR and LPIPS metrics, showing that our method can faithfully reconstruct accurate and realistic rendering.
NeuralBody and SMPLpix are trained on the training poses only, and fail to generalize to novel views, even if the view deviations are very small from the training frames.
VideoAvatar performs a multi-stage optimization, where the geometry is optimized from silhouettes only, and then a texture optimization step is performed.
Errors from the mesh optimization propagate to the texture, leading to a low quality texture, and subsequently a low quality render.
Our method recovers intricate details like loose clothing, loose pants, hoodies, hair, and skirts (Fig.~\ref{fig:geom}).
More qualitative results are in the Appendix.

\begin{table}[t!]
    \centering
    \small
    \begin{tabular}{lrrrr} \toprule
        \multicolumn{5}{c}{\textbf{FID score} $\downarrow$} \\ \midrule
        \textbf{Method} & \textbf{m3c} & \textbf{m4c} & \textbf{f3c} & \textbf{f4c} \\ \midrule
        \textbf{SMPLpix} &  199.04 & 210.27 & 211.09 & 212.16\\
        \textbf{Neural body} & 402.47 & 357.77 & 328.26 & 358.54 \\
        \textbf{VideoAvatar} & 189.82 & \underline{186.03} & \underline{206.07} & 162.79 \\
        \textbf{AnimNeRF} & \underline{183.03} & 200.73 & 237.79 & \textbf{150.80} \\ 
        \textbf{Ours} & \textbf{178.71} & \textbf{184.26} & \textbf{203.42} & \underline{159.31} \\ \midrule
        \multicolumn{5}{c}{\textbf{VGGFace2} $\uparrow$} \\ \midrule
        \textbf{SMPLpix} & .4808 & .6472 & .6222 & .4853\\ 
        \textbf{NeuralBody} & .3713 & .3743 & .4301 & .0000 \\
        \textbf{VideoAvatar} & .8135 & .8799 & .8976 & .8417 \\ 
        \textbf{AnimNeRF} & \textbf{.9079} & \textbf{.8974} & \textbf{.9452} & \textbf{.9259} \\ 
        \textbf{Ours} & \underline{.8766} & \underline{.8926} & \underline{.9380} & \underline{.8948} \\ 
        \bottomrule
    \end{tabular}
    \caption{Quantitative analysis of texture quality of novel poses.} 
    \label{tab:fid}
\end{table}

\subsection{Novel pose synthesis}
\label{sec:novelpose}
We use a set of held-back frames for novel pose synthesis in the ZJU dataset.
Results are in Tab.~\ref{tab:novelview}.
SMPLpix and Neural Body do not generalize well because novel poses \sidedit{that are} not seen during training \sidedit{results in} a distribution shift in the inputs of the frameworks.
VideoAvatar has an uncanny valley effect in the faces of its rendered outputs.
HumanNeRF achieves highly realistic results capturing nuances in body geometry and texture.

\textbf{OOD pose rendering}: However, we notice that the pose distribution is not very different from those in training frames.
Moreover, the people-snapshot dataset doesn't contain frames with other poses than the A-pose.
Therefore, we also compare the realism of textures and faces in an a set of OOD poses.
We curate a set of poses from the AMASS dataset~\cite{AMASS}.
We compare the realism of the models by evaluating the \brandonedit{Fr\'{e}chet Inception Distance (FID score)}~\cite{Seitzer2020FID} of the input frames with novel pose renders, and comparing the face texture of the rendered images with that of the input frames.
Since we use the same novel poses for all methods, the differences in FID must come from texture quality.
To evaluate texture quality of faces, we use face identification as a proxy task.
We use MTCNN \cite{zhang2016mtcnn} to detect faces from images and VGGFace2 \cite{cao2018vggface2} to generate a template feature vector for each method.
We use the face similarity metric between the template of the method and that of the input data, as proposed in \cite{cao2018vggface2}.
Results in Tab.~\ref{tab:fid} and Fig.~\ref{fig:novelpose} shows that our method preserves the subject identity significantly more than VideoAvatar, showing that our method can recover accurate texture 
with a mesh.
AnimNeRF reconstructs the texture well in the parts on the surface, but introduces cloud and tearing artifacts, especially when regions around the unseen areas (armpits and thighs) are stretched too much.
Our method doesn't have such an issue, since our representation is based on a mesh.
Moreover, the texture distortion for novel poses is virtually non-existent.
More qualitative results are shown in Appendix.

\begin{figure}[t!]
    \centering
    \includegraphics[width=\linewidth]{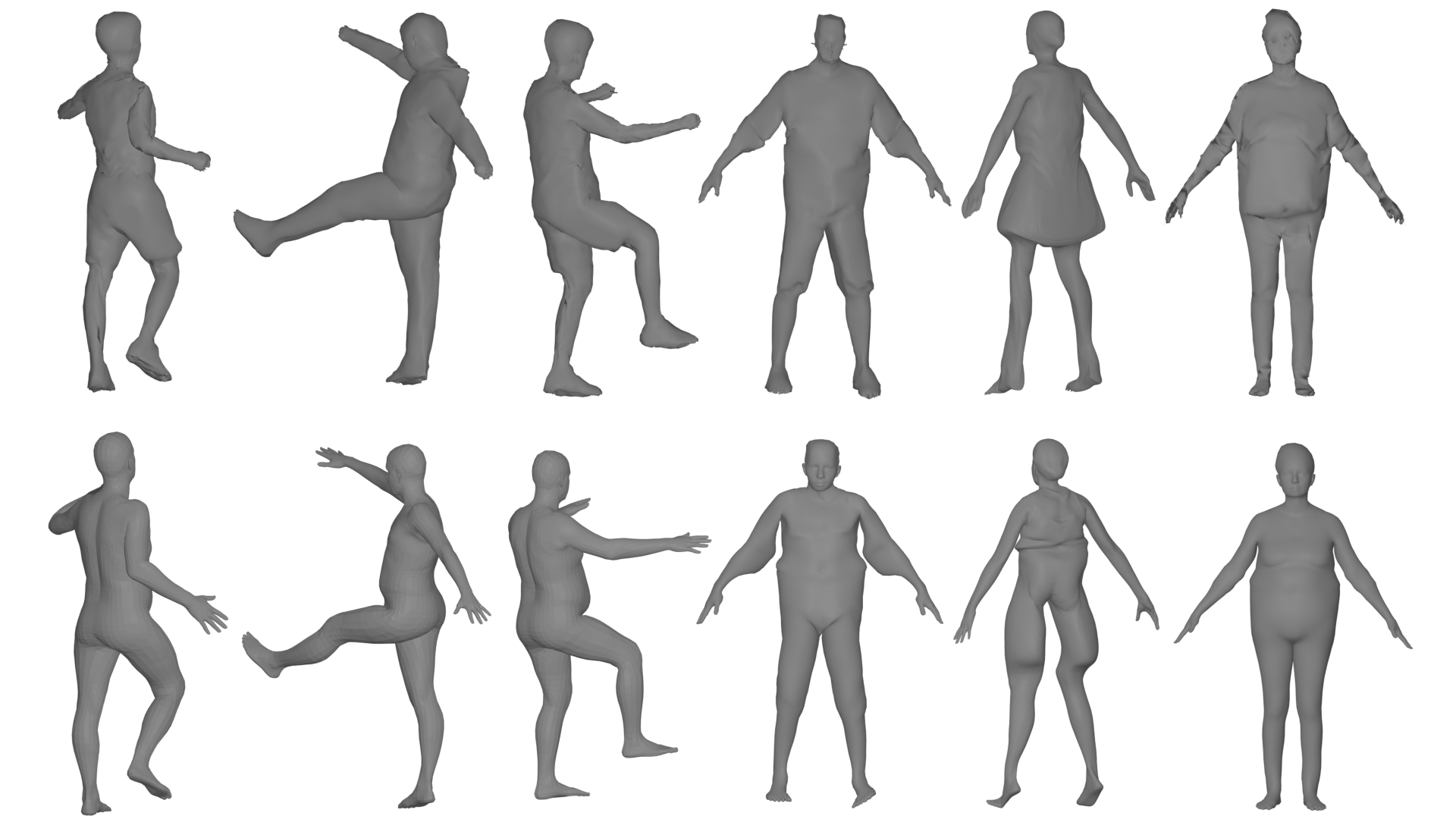}
    \caption{Geometry reconstruction quality of our method (top) and VideoAvatar (bottom) on ZJU Mocap, people-snapshot, and Self-Recon datasets. 
    Our method captures loose fitting pants (1,3) and clothes (2,4,6), hoodie (2), hair (1,2,3), skirts (6).
    }
    \label{fig:geom}
\end{figure}

\subsection{Training/inference time and compute}
\label{sec:trainingtime}

\begin{table}[ht]
    \centering
    \small
    \begin{tabular}{lrrr} \toprule
        \textbf{Method} & \textbf{Training} & \textbf{Inference} & \textbf{GPU usage} \\ 
        & \textbf{time (min)} & \textbf{time (sec)} & (\textbf{GB-hrs}) \\ \midrule
        HN~\cite{weng:cvpr2022:humannerf} & 1013.35 & 3.51 & 399.09 \\ 
        AN~\cite{chen:arxiv2021:animner} & 769.26 & 11.54 & 591.38 \\
        NB~\cite{peng:cvpr2021:neuralbod} & 1020.27 & 0.77 & 62.22 \\
        Ours & \textbf{43.31} & \textbf{0.06} & \textbf{4.11} \\
        \bottomrule
    \end{tabular}
    \caption{Averaged total training and per image inference time (in minutes) and total GPU usage (GB-hr). 
    Our model train upto 24x faster than HumanNeRF on the same data while using upto 4x less compute.
    Results for individual subjects are in the Appendix.
    }
    \label{tab:trainingtime}
    \vspace*{-7pt}
\end{table}

We compare the training and inference time and compute required our method against NeRF methods in Tab.~\ref{tab:trainingtime}.
NeRFs have achieved huge successes in representing scenes faithfully with very accurate rendering.
However, they take a prohibitively long time to train a single scene.
Although several 
improvements have been proposed for static scenes, 
the main bottleneck for AnimNeRF is the KNN step for each sample along the ray, and unposing the transformation to the canonical space.
This is a computationally expensive step since it has to be done for each point from each ray independently.
Moreover, volume rendering leads to a significantly higher inference time and compute requirements ~\cite{Liu21neuralactor}.
\brandonedit{In contrast, unposing is trivial using our method} 
because the rasterization consists of the face index with barycentric coordinates.
\begin{table}[t!]
    \centering
    \small
    \begin{tabular}{llrrrrr} \toprule
    \textbf{Metric} & \textbf{Method} & \textbf{F1} & \textbf{F2} & \textbf{F3} & \textbf{M1} & \textbf{M2} \\
    \midrule
        \multirow{2}{*}{Chamfer} & {VideoAvatar} & 1.47 & 1.05 & 1.41 & 1.20 & 1.08 \\
     & {Ours} & \textbf{1.15} & \textbf{0.96} & \textbf{1.05} &  \textbf{1.01} & \textbf{0.93} \\
     \midrule
    \multirow{2}{*}{P2S} & {VideoAvatar} & 0.59 & 0.46 & 0.57 & 0.54 & 0.45 \\
     & {Ours} & \textbf{0.42} & \textbf{0.40} & \textbf{0.37} & \textbf{0.40} & \textbf{0.35} \\
     \bottomrule
    \end{tabular}
    \caption{Reconstruction loss (cm) on Self-Recon synthetic dataset.} 
    \vspace*{-5pt}
    \label{tab:geom}
\end{table}


\subsection{Geometry reconstruction}
\label{sec:geomrecon}
We quantitatively compare the effectiveness of our method to recover the underlying geometry from the set of images 
using the Self-Recon dataset \cite{Jiang22selfrecon}, which consists of renderings of 5 human subjects with their ground truth meshes.
We compute the average Chamfer distance and Point-to-Surface (P2S) measures.
Our comparison with VideoAvatar~\cite{alldieck:cvpr2018:peoplesnap}, the other mesh-based optimization method is shown in Tab.~\ref{tab:geom}.
Note that our lower distances show that even a low dimensional mesh can capture complex details like loose clothing, hair, etc. with the right optimization and training scheme.   
Qualitative results on all three datasets are in Fig.~\ref{fig:geom} and Appendix.


\section{Conclusion}
\label{sec:conclusion}
There are myriad applications for detailed, personalized, and animatable 3D human models, and they become increasingly practical as generation times, data acquisition, and hardware requirements decrease. 
One promising direction is to bootstrap the training process of human-specific NeRFs ~\cite{weng:cvpr2022:humannerf,chen:arxiv2021:animner} with the geometry and texture learnt from our model.
Since our model directly provides 3D coordinates and RGBs in the canonical space, volume rendering is not needed during this pretraining step.
While NeRFs have demonstrated great versatility, \ie, they are scene-agnostic, 
they are prohibitive for reconstructing well-defined objects with strong priors (\eg faces, human avatars) given their steep training requirements. 
For these applications, we argue that our mesh-based system is competitive and, in some cases, favorable. 
We have demonstrated that our approach is capable of generating results with 
very competitive performance 
to SOTA human-specific NeRFs \cite{chen:arxiv2021:animner,weng:cvpr2022:humannerf}, \brandonedit{but in a tiny} fraction of the training time and compute. 

\clearpage
\bibliography{main}
\bibliographystyle{abbrv}


\end{document}